# Progressive Cross-modal Knowledge Distillation for Human Action Recognition

Jianyuan Ni
j_n317@txstate.edu
Texas State University
San Marcos, TX, USA

Anne H.H. Ngu
angu@txstate.edu
Texas State University
San Marcos, TX, USA

Yan Yan*
yyan34@iit.edu
Illinois Institute of Technology
Chicago, IL, USA

## ABSTRACT

Wearable sensor-based **H**uman **A**ction **R**ecognition (HAR) has achieved remarkable success recently. However, the accuracy performance of wearable sensor-based HAR is still far behind the ones from the visual modalities-based system (*i.e.,* RGB video, skeleton and depth). Diverse input modalities can provide complementary cues and thus improve the accuracy performance of HAR, but how to take advantage of multi-modal data on wearable sensor-based HAR has rarely been explored. Currently, wearable devices, *i.e.,* smartwatches, can only capture limited kinds of non-visual modality data. This hinders the multi-modal HAR association as it is unable to simultaneously use both visual and non-visual modality data. Another major challenge lies in how to efficiently utilize multi-modal data on wearable devices with their limited computation resources. In this work, we propose a novel **P**rogressive **S**keleton-to-sensor **K**nowledge **D**istillation (PSKD) model which utilizes only time-series data, *i.e.,* accelerometer data, from a smartwatch for solving the wearable sensor-based HAR problem. Specifically, we construct multiple teacher models using data from both teacher (human skeleton sequence) and student (time-series accelerometer data) modalities. In addition, we propose an effective progressive learning scheme to eliminate the performance gap between teacher and student models. We also designed a novel loss function called **A**daptive-**C**onfidence **S**emantic (ACS), to allow the student model to adaptively select either one of the teacher models or the ground-truth label it needs to mimic. To demonstrate the effectiveness of our proposed PSKD method, we conduct extensive experiments on Berkeley-MHAD, UTD-MHAD and MMAct datasets. The results confirm that the proposed PSKD method has competitive performance compared to the previous mono sensor-based HAR methods.

---

*Corresponding Author.

## 1 INTRODUCTION

**H**uman **A**ctivity **R**ecognition (HAR) is an active research area due to its widespread applications, for example in human-robot interaction and in smart health area [78]. There are currently two mainstreams of HAR systems: namely, visual modalities-based (*i.e.,* RGB, skeleton and depth) and non-visual modalities-based systems (*i.e.,* audio, accelerometer data, WiFi, and RFID) [70]. Among visual modalities-based systems, despite the significant achievements of video-based HAR methods had made, privacy concerns in video/image data has drawn increasing attentions recently [11, 55, 69, 94]. For instance, one of the most influential datasets in the computer vision area, ImageNet, has released an updated version that blurs people's faces for privacy protection [94]. Consequently, pure video-based approach is infeasible to be used in privacy-sensitive areas. Instead, skeleton modality can eliminate the privacy concerns while encoding the trajectories of human body joints to characterize the geometric 3D body movement patterns in a continuous way [33]. Also, skeleton modality is not susceptible to background variations and thus has attracted a lot of attentions recently [33, 93]. However, such skeleton-based systems, as well as other visual modalities-based systems, fail to be practical for real-time HAR, or anytime and anywhere HAR monitoring applications.

On the other hand, thanks to the development of Internet of Things (IoT) devices, time-series data from wearable devices has provided new opportunities in solving sensor-based HAR problem [45, 47–49, 56, 62, 80]. Currently, one of the most common sensors used in HAR problem is the accelerometer data due to its small footprint and being available on many low cost sensor devices[11]. However, the accuracy performance of sensor-based HAR system is far behind when compared to the video-based HAR system as RGB video contains richer information and can capture scene context [70]. For example, previous work demonstrated that, by using accelerometer data from a wrist-worn watch, the deep learning method for fall detection only reach 86% accuracy performance [45]. This is because the constraint of a single context from the accelerometer data lack the 3D information and can not discriminate various wrist movements when someone falls [20].

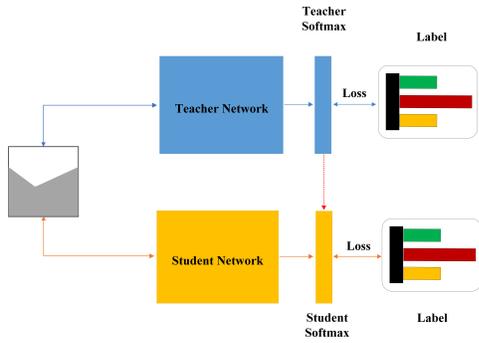

(a) **Conventional two-steps knowledge distillation method with pre-trained teacher model [21, 61, 98].**

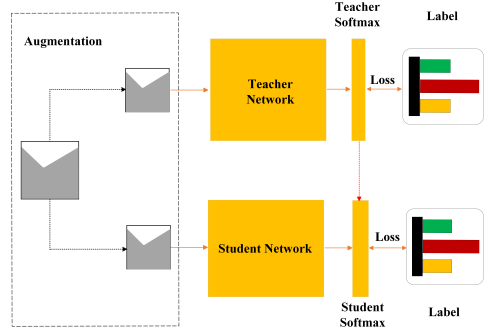

(b) **One-step self-distillation via augmentation [28, 92].**

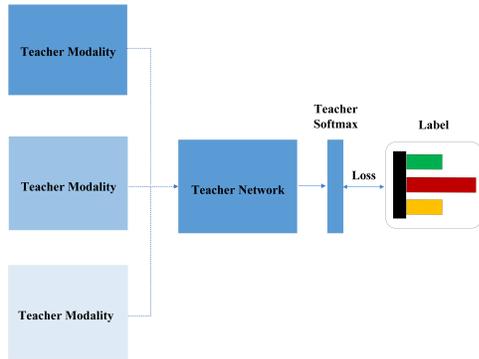

(c) **Cross-model distillation with multi-teacher models using various modality data [42].**

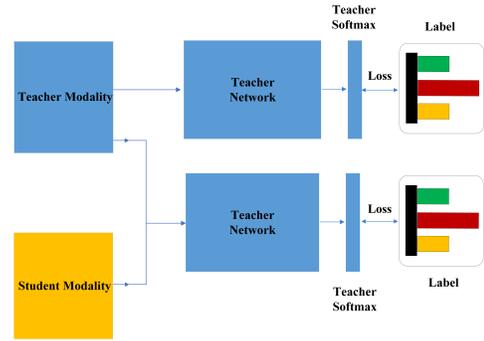

(d) **Proposed multi-teacher method. One teacher model took teacher modality only as the input, while the other one use modality data from both teacher and student domain.**

**Figure 1: Comparison of various distillation methods.**

In real life, humans perceive the surrounding world in a multimodal cognition way. Similarly, multi-modal machine learning is a modeling approach trying to learn complementary features from diverse modalities of data. As a result, multi-modal approaches can often lead to more robust algorithms and better HAR performances. However, current wearable devices can only acquire certain kinds of non-visual modality data, such as accelerometer and gyroscope data [77]. This prevents the multi-modal HAR implementation on the wearable devices as it fails to use both visual and non-visual modality data simultaneously. Also, such multi-modal methods usually have complex architectures and incur high computational overheads which wearable devices can not afford. For example, previous study indicated that a smartphone (LG Nexus 5X, 1.8 GHz, Hexa-core processor with 2G of RAM) can only support a long short-term memory (LSTM) model which contains an input layer, two hidden layers, and an output layer [44].

How to leverage the advantages of advanced multi-modal methods for the wearable sensor-based HAR problem? The technique of cross-modal transfer, *i.e.,* knowledge distillation, can be one of the potential solutions. Knowledge distillation (KD) was formally popularized by distilling knowledge from a larger model (*i.e.,* teacher) into a smaller model (*i.e.,* student) as a two-steps process as shown in Figure 1a. By mimicking the pre-trained teacher model, the student model is able to retain similar accuracy as well as reducing the computation resource demand [21]. After that, the data augmentation based self-knowledge distillation methods [28, 92, 97], which adopted a consistent prediction of relevant data from the same class, *i.e.,* distorted versions of instances, has proposed to improve the performance of the student model as shown in Figure 1b. For example, Zhang *et al.* proposed a one-step self-distillation method, in which knowledge from the deeper parts of the network is distilled into its shallow sections [101]. Currently, there are only a few multi-modal KD approaches for the HAR problem [30, 42, 52, 72]. For example, Liu *et al.* [42] introduced a multi-modal KD method which integrated various sensor information to improve the vision modality as shown in Figure 1c. Instead, Ni *et al.* [52] proposed a multi-modal KD approach where the complementary information from the video domain was adaptively transferred to the sensor domain. Although those studies provide promising results on multi-modal HAR problem, there are two questions that those works have not addressed: 1) they have used a pre-trained teacher model to directly guide the student network, which is an inefficient learning process and thus contributed to the student model's performance degradation; 2) they failed to consider the fact that the student modality can also contribute to the whole KD process. We argued that there are two key factors in the KD process: knowledge source (*i.e.*, *teacher model*) and the distillation process. In order to increase the accuracy performance of a student model, a teacher model with higher performance (*i.e.*, *stronger teacher*) should be achieved at first. In addition, how to reduce the performance gap between teacher and student model via novel training strategy is another important step to produce a strong student model.

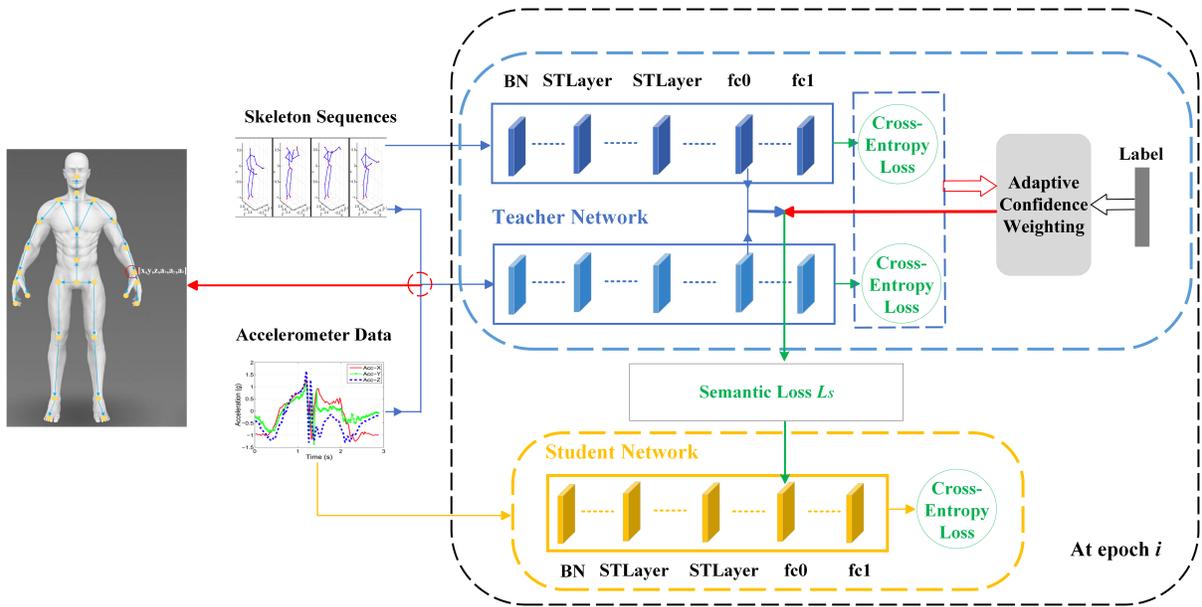

Figure 2: Schematic overview of the proposed PSKD method. Initially, multi-teacher models are constructed using both teacher and student modalities. Next, we propose a progressive learning scheme to eliminate the performance gap between teacher and student models. We also introduce a loss function to allow the student model adaptively decide either one of the teacher models or the ground-truth label it needs to mimic.

Driven by the aforementioned two intuitions, we proposed an end-to-end Progressive Skeleton-to-Sensor Knowledge Distillation (PSKD) for HAR recognition in this study. The overview of the proposed method is shown in Figure 2. First, we propose a new multi-teacher approach to construct multiple teacher models using skeleton (teacher) and accelerometer (student) data modalities as shown in Figure 1d. In this way, the teacher models can also understand the characteristic of the student modality data so that teacher models can generate models which are easier for student model to mimic. Next, we design an effective progressive learning (PL) scheme to eliminate the performance gap between teacher and student models. Specially, the student model will be updated after the multi-teacher models are updated every epoch to converge to the ground-truth labels. During the PL training process, a novel loss function called Adaptive-Confidence Semantic (ACS), is introduced to allow the student model adaptively decide which teacher models or the ground-truth label it needs to mimic. In summary, the contributions of this paper are summarized as follows: 1) To the best of our knowledge, this is the first study conducting the cross-modal KD model from the skeleton data domain to the wearable sensor data domain. In this PSKD model, a student model with input of accelerometer data, learns the compensatory information from multi-teacher models with both input of skeleton sequences and wearable sensor data. 2) We designed an effective PL scheme coupled with a novel loss function (ACS), which is utilized to alleviate the modality gap between the teacher and the student model. 3) We demonstrated the competitiveness performance of the proposed PSKD method on three public datasets over the previous sensor-based HAR methods.

## 2 RELATED WORK

We briefly review the existing studies for the skeleton and wearable sensor-based HAR problem. Multi-modal HAR and knowledge distillation work are also included in this section.

### 2.1 Skeleton-based HAR

Skeleton sequences encode the trajectories of human body joints, which characterize temporal contextual informative human motions over time. There are several advantages of using skeleton sequences for HAR problem, due to its informative representation and its robustness against variations of backgrounds [70]. Early skeleton-based HAR works mainly focused on extracting hand-crafted spatial and temporal features, which can be divided into joint-based [79] and body part-based [76] methods. Besides, due to the strong feature learning capability, skeleton-based deep learning methods have become one of the mainstream research in this field. Recurrent neural network (RNN) or their variants (*e.g.*, long short-term memory (LSTM)) are capable of learning the dynamic dependencies in sequential data [12, 38, 102, 103]. For example, Du *et al.* introduced an end-to-end RNN, which divided skeleton data into five body parts rather than taking the skeleton from each frame as a whole. These five body parts were then fed to several bidirectional RNNs to generate high-level representations of the action [12]. Liu *et al.* proposed the tree structure-based skeleton traversal method to exploit the spatial information of the skeleton sequences [37]. Liu *et al.* presented an attention-based LSTM network to encode the skeleton sequence and refine the global context on the informative joints [39]. Similarly, a deep LSTM network consisted of the jointly spatial and temporal attention subnetwork was proposed to model the temporal dynamics and spatial configurations [67].

At the same time, due to the expressive power of graph structures, Graph convolutional Networks (GCNs) has been introduced to the skeleton-based HAR problem [51, 93, 104]. For example, Yan *et al.* exploited GCNs for skeleton-based HAR by introducing Spatial-Temporal GCNs (ST-GCNs) that can automatically learn both the spatial and temporal patterns for skeleton-based HAR problem [93]. Wu *et al.* adopted the ST-GCN to learn the global information at first and then designed a residual layer to capture the spatio-temporal information of skeleton sequences [87]. Li *et al.* proposed a spatial and temporal graph router to produce new skeleton-joint-connectivity graphs. After that, this skeleton-joint-connectivity graphs was fed to the ST-GCN for further classification [33]. Moreover, Shi *et al.* proposed a two-stream Adaptive GCN (2s-AGCN), which coupled the first-order information (coordinates of joints) with the second-order information (lengths and directions of human bones) on the skeleton-based HAR [63]. Li *et al.* proposed the Symbiotic Graph Neural Networks (Sym-GCNs) to handle both action recognition and motion prediction tasks simultaneously so that these two tasks can enhance each other [35]. In order to reduce computational costs of GCNs, Cheng *et al.* present a ShiftGCNs which applied lightweight point-wise convolutions and shift graph operations [7]. Similarly, Song *et al.* proposed a multi-stream GCNs model which fuses joint positions, motion velocities as well as bone features. Separable convolutional layers and compound scaling strategy was applied in this study to reduce the redundant training parameters [68]. In summary, the skeleton modality provides the body structure information, which is effective for HAR problem. Nevertheless, skeleton-based approaches, as well as other visual modalities-based approaches, can only be applied in a static environment where visual devices can be permanently installed. For example, visual modalities-based approaches are not suitable for outdoor HAR monitoring problem.

## 2.2 Wearable sensor-based HAR

Wearable sensor-based HAR methods has received huge attentions due to their robustness against occlusion and viewpoint variations [70]. Wearable sensors only includes subtle intra-class variations for the same action performance, regardless of the size of human body which varies from person to person. Therefore, wearable devices has been adopted for remote monitoring systems without worry about the privacy-safety concerns [27, 46]. Numerous Convolutional neural network (CNN) on wearable-based HAR [6, 32, 99] has been proposed. For instance, a wrist worn tri-axial accelerometer was used to perform arm movement prediction and results demonstrated the robustness of such approach [56]. RNN type of model was also suggested to deal with time-dependent input sequences [44, 45, 54, 89]. Wang *et al.* [80] integrated a CNN and bidirectional

LSTM model to acquire spatial and temporal features from acceleration data. Zhao *et al.* proposed the residual bi-directional LSTM model to concatenate the forward and backward state *i.e.,* positive and negative time direction to avoids the gradient vanishing problem [106]. Wang and Liu [82] present a novel Hierarchical LSTM method to improve the system's performance. Meanwhile, some approaches also suggested converting wearable sensor sequences as images for HAR study. Zeng *et al.* [99] transformed the single-axis sensor data into one-dimensional images and then fed them to CNN for identification. Lu *et al.* [43] encoded the tri-axial acceleration data into color images, which were fed into a ResNet for HAR. However, the accuracy performance of sensor-based HAR is still far behind compared to the visual modalities-based HAR results due to the constraint of a single contextual information from accelerometer data [20]. In reality, we human, understand the surrounding environment in a multi-modal way. Hence, by utilizing the complementary information acquired from different modalities, it is possible to enrich the gained knowledge and thus enhance the sensor-based HAR performance eventually.

## 2.3 Multi-modal based HAR

Recently, deep learning methods have been conducted on HAR problem [1, 2, 9, 10, 59, 86]. For example, Dawar *et al.* proposed the data augmentation CNN and CNN+LSTM methods based on depth and inertial modalities, respectively. Wei *et al.* fed the 3D videos frames as well as 2D inertial images to a 3D CNN and a 2D CNN models, respectively. The score fusion strategy outperformed the feature fusion method in this study [86]. Similarly, some studies have also been conducted on the depth-inertial fusion techniques by combining two-stream CNN architectures [1, 2]. Besides that, Islam and Iqbal [24] also proposed a separate encoder to fuse RGB,skeleton and inertial modalities in a similar shaped vector representation way. Li *et al.* adopted the ST-GCN model [93] to extract the skeleton feature vector from videos and the R(2+1)D [34] model to encode the RGB videos directly. While the aforementioned multi-modal models tend to achieve better performance, one of the drawback is the high computational overhead and larger memory demand. Consequently, efficient model compression methods have emerged to build deep models with less computational resource and maintain the similar performance [19].

Knowledge distillation (KD) is one of the model compression methods which transfer knowledge from a computational expensive model into a smaller network [21]. In general, student model tried to mimic the performance from a pre-trained teacher model as shown in Figure 1a. Zagoruyko and Komodakis [98] proposed the attention information transfer method by forcing a student CNN model to mimic the attention maps from a teacher network. Park *et al.* designed the distance-wise and angle-wise distillation loss for the relational knowledge transfer in the KD process [57]. Tung *et al.* proposed a new form of KD distillation loss with the constraint that input pairs that produce similar activations in the teacher network should also produce similar activations in the student network [73]. Different from these KD methods that mainly focus on the distillation loss task, there are only a few multi-modal KD approaches for the HAR problem [16, 22, 30, 42, 52, 72]. For example, Hoffman *et al.* designed a modality hallucination architecture by using depth as side information to guide an RGB object detection model [22]. Garcia *et al.* built a KD framework to learn representations from the depth and RGB videos, while only use RGB data at test time. Similarly, Thoker *et al.* proposed a multi-modal KD framework which used RGB videos to train the teacher network for HAR task. After that, two student networks were trained using mutual learning to improve the performance [72]. In addition, Ni *et al.* present the first multi-modal KD approach on the sensor-based HAR problem. In this study, the complementary information from the video domain was adaptively transferred to the sensor domain and improve the accuracy performance of sensor-based HAR problem [52]. However, previous works either ignored the fact that the student modality can contribute useful information to the training of the KD process or it is not efficient to let the small student network learn directly from a pre-trained teacher model. Our work, instead, utilized both teacher and student modality data to build up multiple teacher models. In this way, teacher models will tend to produce a model which is easier for the student model to understand from the human's learning analogy perspective.

## 3 METHODS

This section describes our proposed approach in terms of the multi-teacher models construction process, the progressive learning KD procedure, and the designed loss function used to let the student model adaptively learn from either one of these multiple teacher models or the ground-truth label directly.

### 3.1 Multi-teacher Construction Process

While impressive progress has been achieved under the standard teacher-student KD paradigm, the intuition that a student can learn more effectively from multiple teachers has only been investigated recently. Currently, there are several studies using multiple teacher models in the KD process [17, 40, 42, 50, 71, 81, 88, 90, 95]. However, these works ignore the following two advantages of using the teacher model that is familiar with the student's modality data: 1) when the teacher model realizes and understands the intra-modality characteristics between teacher and student domain modality data, it can generate a model which alleviate some level of difficulties when the student model tries to mimic its performance; 2) additional modality data, from both teacher and student domain, can build up teacher models with better performance.

Based on the above two perceived advantages, we constructed our multi-teacher models which use input modality from both teacher (*i.e.,* skeleton sequence) and student domain (*i.e.,* accelerometer data). Inspired by [14], we use the well established Graph Convolutional Networks (GCNs) models, Adaptive GCNs [63], to utilize the skeleton sequences as the input for the first teacher model $Teacher_{sk}$. After that, we concatenated the skeleton sequence and accelerometer data together to build another teacher model $Teacher_{fu}$ as shown in Figure 3. We briefly introduce the fusion process here: let $\mathbf{X}_{SK} \in \mathbb{R}^{(M, C_{SK}, T_{SK}, N_{SK})}$ be a skeleton sequence input, where $M$ is the number of participants that are involved in an action, $C_{SK}$ is the initial 2D joint coordinates and size $T_{SK}$ and $N_{SK}$ are the sequence length and number of skeleton graph nodes. For accelerometer data, the input is defined as $\mathbf{X}_{AC} \in \mathbb{R}^{(M, C_{AC}, S_{AC}, T_{AC})}$, where $T_{AC}$ is the accelerometer se-quence length, $S_{AC}$ is the number of sensors and $C_{AC}$ is the channel dimension of the accelerometer data. For example, given the accelerometer data from a smartwatch with x-, y- and z-values, the structure would be $S_{AC} = 1$ and $C_{AC} = 3$. Similar to the skeleton data, $M$ denotes the person wearing the sensor device and $C_{AC} = C_{SK} = 3$ since skeleton sequences and accelerometer data all include x-, y- and z-axis values. As a result, if there is only one participant wearing a smartwatch during the activity performance process, the number of $M_{SK}$ should be equivalent to that of $M_{AC}$. Also, a common $T$ can be guaranteed by resampling $T_{SK}$ and $T_{AC}$ to the same time length. After that, the fused data which formed as $\mathbf{X} \in \mathbb{R}^{(M, C_{SK} + C_{AC}, T, N_{SK})}$, can be fed into the AGCN backbone as shown in Figure 2.

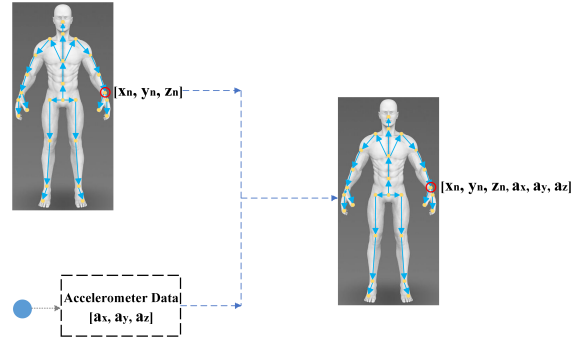

**Figure 3: The fusion of skeleton sequence $[x_n, y_n, z_n]$ and accelerometer data $[a_x, a_y, a_z]$.**

### 3.2 Progressive Learning KD Procedure

Hilton *et al.* [21] proposed the standard KD process which referred to a model-agnostic method where a small or less complex model (*i.e., student*) tried to minimize the statistical discrepancy between its predictions distributions and the predictions of a complicated model (*i.e., teacher*). However, the standard KD process cannot fully solve the performance gap between various modalities [42, 52]. We face the problem that the distribution of teacher and student modality could be very far from each other at the beginning, leading to a difficult distillation process. It has been proved that a more powerful teacher model is not a guarantee to give rise to a better student model during the KD process [60]. Also, it is hard to let the student model simply mimic a pre-trained teacher model when the capacity gap between the teacher and student is large [25, 50].

In order to solve this problem, an intermediate network has been introduced recently [50, 60, 64]. For instance, Mirzadeh *et al.* proposed the Teacher-Assistant KD method by gradually increasing the teacher size to foster the distillation process [50]. This is mainly due to the fact that the learning process can be evaluated if a suitable goal of the teacher model is set for the student model to follow. However, training intermediate networks will incur more computational cost and training time [60, 64]. Therefore, different to the original KD process where the student learns directly from the pre-trained teacher model, inspired by [64], we introduced a progressive KD learning procedure that requires the student model to learn from the teacher model step-by-step or incremental fashion which can help the student model better mimic the performance of the teacher model. Specially, we update the student model immediately after the teacher model updates one step towards the ground-truth labels.

However, with multiple teacher models in this study it can be confusing for the student model to mimic their performances when their prediction results are inconsistent or incorrect as shown in Figure 4. For example, when both teacher models ($Teacher_{sk}$ and $Teacher_{fu}$) predict incorrectly or when only one of the teacher models ($Teacher_{sk}$ or $Teacher_{fu}$) achieve the correct prediction result, it is unrealistic for the student model to imitate their performances directly. Technically, the student model should have the knowledge to select which teacher model or the ground-truth label it needs to follow adaptively.

### 3.3 Adaptive-Confidence Loss Function

Currently, several multi-teacher model studies are proposed and the results demonstrated the beneficial effect of multi-teacher models on the KD process [15, 31, 88, 96]. For example, either fixed weight assignment [15, 88, 96] or other label-free schemes, such as entropy-based weight optimization method [31], has been used for the student model to learn from the multi-teacher models. However, fixed weight assignment failed to balance the importance of multi-teacher models and the other methods may misguide the student model in the presence of low-quality teacher predictions. More recently, Zhang *et al.* proposed the confidence-aware loss function which adaptively assign the sample-wise reliability for each teacher prediction based on the ground-truth labels [100]. However, they failed to realize the case where all teacher models predicted wrong. Based on these observations, we proposed a novel Adaptive-Confidence loss (AC) to let the student model adaptively emulates the best teacher model performance or the ground-truth label during the KD process. In this study, there are two teacher models, leading to four different cases we need to consider to design the AC loss intuitively: 1) when both teacher models, $Teacher_{sk}$ and $Teacher_{fu}$, all have correct prediction results, the student model should mimic the teacher model which achieved the higher prediction score; 2) when only one of the teacher models predict correctly, the student model shall mimic the one which predict correctly; 3) when both teacher models predict incorrectly, the student model will have to switch and simulate the ground-truth label instead. An illustration example of these fours cases is shown in Figure 4.

Given a teacher model $T_k$ and a student model $S_k$, the soft-target $\tilde{y}^T$ produced by the teacher model is considered as high-level knowledge. The loss of KD when training a student model can be defined as:

$$\mathcal{L}_{KD} = \mathcal{L}_C(y, y^S) + \alpha\mathcal{L}_K(\tilde{y}^T, \tilde{y}^S) \tag{1}$$

$$\mathcal{L}_K = \frac{1}{m}\sum_{k=0}^{m} KL(\frac{P^{T_k}}{T}, \frac{P^{S_k}}{T}) \tag{2}$$

where $y$ and $y^S$ refer to the predicted labels and class probability for the student network in this study, respectively. $\tilde{y}^S$ is the soft target generated by the student model. Here $\mathcal{L}_C$ is the typical cross-entropy loss and $\mathcal{L}_K$ is the Kullback-Leibler (KL) divergence, while $P^{T_k}$ is the class probability for the teacher network and $P^{S_k}$ is the class probability for the student network. $T$ represents the temperature controlling the distribution of the provability and we use $T = 4$ in this study according to [21].

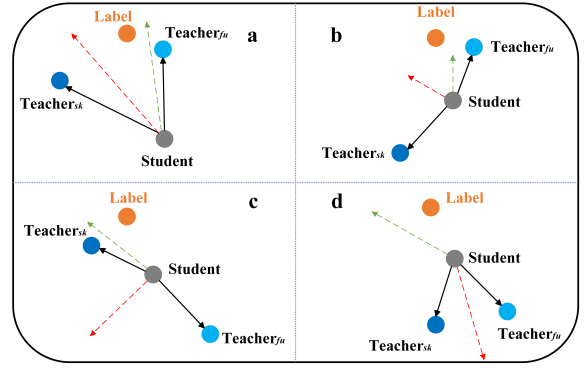

**Figure 4: Comparison of the previous average weight result (red dash line) and our proposed Adaptive-Confidence weight method (green dash line).**

Intuitively, we assign different weights on each teacher model by calculating the cross entropy loss between teacher predictions and the ground-truth labels:

$$\mathcal{L}_{CE_{KD}}^K = -\sum_{c=1}^{C} y^c log(\alpha(Z_{T_K}^c)), \tag{3}$$

$$\omega_{KD}^k = \begin{cases} \frac{1}{K-1}\left(1 - \frac{exp(L_{CE_{KD}}^k)}{\sum_j exp(L_{CE_{KD}}^j)}\right), & \mathcal{L}_{CE_{KD}}^K \leq \mathcal{L}_{KD} \\ 0, & \mathcal{L}_{CE_{KD}}^K \geq \mathcal{L}_{KD} \end{cases} \tag{4}$$

where $T_k$ denotes the $k$th teacher and $\alpha(z^c)$ is the softmax function. Consequently, the overall teacher predictions are then aggregated with calculated weights:

$$\mathcal{L}_{MK} = -\sum_{k=1}^{K}\omega_{KD}^k \sum_{c=1}^{C} Z_{T_K}^c log(\alpha(Z_S^c)), \tag{5}$$

Therefore, the teacher model whose prediction is closer to the ground-truth labels will be assigned larger weight $\omega_{KD}^k$. In addition, when all of the multi-teacher model predict incorrectly, the student model will try to follow the ground-truth label instead.

Since multi-modal action data share the same semantic content [42, 52], semantic loss is defined as:

$$\mathcal{L}_S = \frac{1}{m}\sum_{k=1}^{m}(\|H^S - H^T\|)_2^2 \tag{6}$$

where $H^S$ and $H^T$ represent the feature of fc0 layer from both student and teacher models. To keep $H^S$ and $H^T$ spatial dimensions same, we add one more fc layer (fc0) before its original fc layer (fc1) shown in Figure 2.

In summary, we use the original KD loss $L_{KD}$ and augment it to include Adaptive-Confidence loss $L_{MK}$ as well as the semantic loss $L_S$, to train the student network and the final ACS loss for the student model is defined as follow:

$$\mathcal{L} = L_{KD} + \beta L_{MK} + \gamma L_S \tag{7}$$

where $\alpha, \beta, \gamma$ are the tunable hyperparameters to balance the loss terms for the student network.

## 4 EXPERIMENTS

### 4.1 Dataset

In this study, three benchmark datasets were selected due to their multi-modal action forms:

*Berkeley-MHAD* [53]. This dataset includes 11 action classes performed by 12 participants (5 females and 7 males). There are 6 three-axis wireless accelerometers installed to measure movement at the wrists, ankles and hips. In summary, there are 3,948 accelerator samples in total. We use skeleton motion data and accelerometer data as the teacher modality and accelerator data as the student modality. In this study, we use the first 7 participants for training and the rest ones for testing mentioned in [53].

*UTD-MHAD* [5]. This dataset covers 27 action classes performed by 8 participants (4 females and 4 males). In this study, we use skeleton sequences and inertial data (accelerometer data) as the teacher modality and accelerometer data as the student modality. Both modalities have 861 samples and we spit them in half for training and testing mentioned in [5].

*MMAct* [30]. This dataset includes 37 action classes performed by 20 participants (10 females and 10 males) containing more than 36,000 trimmed clips. Since the skeleton sequences are missing in this dataset, we use OpenPose to extract them from RGB videos [4]. After that, we use skeleton sequences plus accelerometer data from watch as the teacher modality and accelerometer data from watch as the student modality. One of the various settings (cross-subject) is used to evaluate this dataset based on the train and test split strategy mentioned in [30].

### 4.2 Experimental Settings

All the experiments were performed on four Nvidia GeForce GTX 1080 Ti GPUs using PyTorch. To guarantee a deterministic and reproducible behavior, all training procedures are initialized with a fixed random seed. We employed the classification accuracy and F-measure as the evaluation metric to compare the performance of the PSKD model. Grid-search method [42] was conducted to evaluate the effect of hyper-parameters $\alpha, \beta, \gamma$ in three datasets.

### 4.3 Comparison to the State-of-the-Art

We compare the performance of our PSKD with state-of-the-art vision-based action recognition (VAR), multi-modal action recognition methods (MMAR), skeleton-based action recognition (SKAR), sensor-based action recognition (SAR), and knowledge distillation (KD) methods. The comparison results of three datasets are shown in Table 1, 2, and 3, respectively. In Table 1, the proposed PSKD model performs better than all the previous comparable VAR models when the RGB videos used as the input data by 0.79%-9.59% [23, 41, 65, 85, 105]. We make an improvement in the testing accuracy of 4.99% compared to the study where 16 features from accelerometer signals were captured for classification [18]. Similarly, the proposed PSKD model achieved higher accuracy performances compared to the previous MMAR and SKAR models [24, 91]. Especially, the proposed PSKD model outperforms all previous SAR methods using both accelerometer and gyroscope data as the input,

Table 1: Comparison results between our proposed method and state-of-the-art methods on UTD-MHAD dataset in accuracy performance (%). Acc. denotes accelerometer and Gyro. denotes gyroscope.

| Type | Method | Testing Modality | Accuracy (%) |
|---|---|---|---|
| VAR | Hussein *et al.* [23] | RGB video | 85.60 |
| | Wang *et al.* [85] | RGB video | 85.81 |
| | Zhao *et al.* [105] | RGB Video | 92.10 |
| | Si *et al.* [65] | RGB Video | 94.40 |
| | Liu *et al.* [41] | RGB Video | 92.84 |
| SKAR | Xiao *et al.* [91] | Skeleton | 94.37 |
| MMAR | Islam *et al.* [24] | Skeleton+RGB video | 95.12 |
| SAR | Singh *et al.* [66] | Acc. + Gyro. | 91.40 |
| | Ahmad and Khan [2] | Acc. + Gyro. | 95.80 |
| | Wei *et al.* [86] | Acc. + Gyro. | 90.30 |
| | Garcia-Ceja *et al.* [18] | Acc. | 90.20 |
| KD | Ni *et al.* [52] | Acc. | **96.97** |
| Proposed | PSKD | Acc. | 95.19 |

Table 2: Comparison results between our proposed method and state-of-the-art methods on Berkeley-MHAD dataset in accuracy performance (%). Acc. denotes accelerometer.

| Type | Method | Testing Modality | Accuracy (%) |
|---|---|---|---|
| VAR | Wang *et al.* [84] | RGB Video | 88.19 |
| | Zhou *et al.* [107] | RGB Video | 95.32 |
| | Lin *et al.* [36] | RGB Video | 96.87 |
| SKAR | Vantigodi *et al.* [74] | Skeleton | 96.06 |
| | Vantigodi *et al.* [75] | Skeleton | 97.58 |
| | Kapsouras *et al.* [26] | Skeleton | **98.18** |
| SAR | Das *et al.* [8] | Acc.(Six locations) | 88.90 |
| KD | Ni *et al.* [52] | Acc. (Left Wrist) | 90.02 |
| Proposed | PSKD | Acc. (Left Wrist) | 94.76 |

Table 3: Comparison results between our proposed method and state-of-the-art methods on MMAct dataset in F1 socre performance (%). Acc. denotes accelerometer and Gyro. denotes gyroscope.

| Type | Method | Testing Modality | Cross Subject(%) |
|---|---|---|---|
| VAR | Kong *et al.* [29] | RGB video | 62.80 |
| | Wang *et al.* [85] | RGB video | 64.40 |
| | Zhou *et al.* [107] | RGB video | 66.56 |
| | Lin *et al.* [36] | RGB video | 70.12 |
| | Kong *et al.* [29] | RGB video | 59.10 |
| SAR | Kong *et al.* [30] | Acc.(watch+phone) | 62.67 |
| KD | Ni *et al.* [52] | Acc. (watch) | 60.14 |
| Proposed | PSKD | Acc. (watch) | **71.42** |

which validate the effectiveness of the PSKD model. However, it is worth noting that the proposed PSKD model does not perform better as compared to the previous method where the accelerometer data learn the complementary information from video domain during the KD process [52]. This degradation was mainly due to the noisy data in the skeleton domain from the UTD-MHAD dataset[13]. In general, these results demonstrated that accelerometer data in the

PSKD model can achieve competitive accuracy performances. In Table 2, even though the proposed PSKD model can not outperform the previous SKAR and VAR models where the skeleton data or RGB video was used during the testing modality [26, 36, 75], our proposed PSKD method tested with only the left wrist accelerometer data does perform better compared to the previous study where accelerometer data from six locations were used [8, 52], regardless of any data prepossessing they applied. This result sheds light on the proposed PSKD for improving sensor-based HAR. In Table 3, while accelerometer data from the watch is the only modality in the testing phase, the method achieves better F-score performance compared to [29, 30, 36, 83, 107] in which either video streams or accelerometer data from phone and watch were used in the testing phase. This result validates that accelerometer data in the PSKD model can significantly learn knowledge from skeleton data and thus effectively improve sensor-based HAR performance.

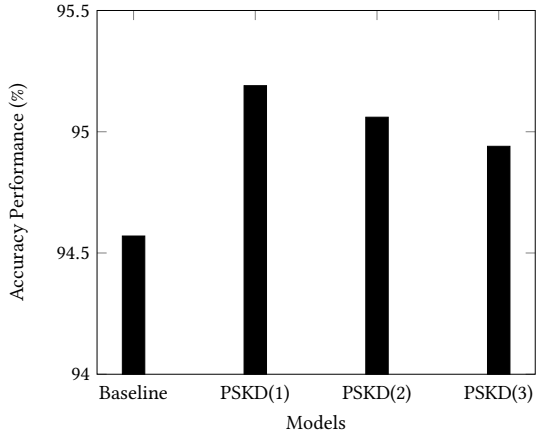

**Figure 5: Accuracy performances (%) between the baseline and the proposed PSKD method on UTD-MHAD dataset [5]. The number in parenthesis means the epoch numbers which the student tries to mimic either the best teacher model or the ground-label truth iteratively.**

## 4.4 Ablation Study

In this subsection, we design experiments to verify the effectiveness of each component in the proposed framework based on UTD-MHAD dataset [5] and try to answer the following questions:

**(i) What is the effectiveness of multi-teachers progressive learning scheme in PSKD?**

To evaluate the effectiveness of the proposed multi-teachers progressive learning (PL) scheme in PSKD method, we compare the PSKD with the student baseline: 1) a student baseline model which learns directly from two pre-trained teacher model ($Teacher_{sk}$ and $Teacher_{fu}$) ; 2) PSKD model with different epoch numbers which the student tries to mimic. As shown in Figure 5, the proposed PSKD model outperforms the student's baseline model by 0.37%-0.62%, proving that the PL scheme is able to reduce the performance gap and thus improve the accuracy performance of the student model. In addition, PSKD(1) achieved higher accuracy performance compared to PSKD(2) and PSKD (3), indicating that a smaller learning step can boost the learning capacity of the student model.

**Table 4: Ablation study of accuracy (%) and F1 score (%) performance on UTD-MHAD dataset. Acc. denotes accelerometer and W/O denotes without. AC denotes the Adaptive-Confidence loss $L_{MK}$. S denotes semantic distillation loss $L_S$.**

| Method | Testing Modality | Accuracy | F1 score |
|--------|------------------|----------|----------|
| Logits [3] | Acc. | 93.87 | 94.15 |
| Fitnet [61] | Acc. | 94.03 | 94.27 |
| ST [21] | Acc. | 94.34 | 94.64 |
| AT [98] | Acc. | 94.27 | 94.80 |
| RKD [57] | Acc. | 95.03 | 95.02 |
| SP [73] | Acc. | 94.06 | 94.56 |
| CC [58] | Acc. | 94.12 | 94.72 |
| **ACS** | Acc. | **95.19** | **95.67** |
| AC(W/O S) | Acc. | 94.51 | 94.59 |
| S (W/O AC) | Acc. | 94.82 | 94.21 |

**(ii) What are the contributions of each loss terms in the proposed ACS loss function?**

To evaluate the effectiveness of the proposed loss function, we compare the ACS function with state-of-the-art KD methods [3, 21, 57, 58, 61, 73, 98]. For those methods, we use the shared codes, and the parameters are selected according to the default setting. As shown in Table 4, the proposed ACS loss function performs better than all of the comparable KD functions. These observations validates that our ACS can effectively transfer the knowledge from skeleton modalities to wearable sensor modalities by integrating two complementary modules, $L_{MK}$ and $L_S$. In addition, the Adaptive-Confidence loss $L_{MK}$ contributes about 0.31% to accuracy improvement as compared to semantic distillation loss $L_S$, which validates the assumption that distillation process is a key factor in the KD process. Also, semantic distillation loss $L_S$ contributes 0.17% to accuracy improvements, proving that the semantic information is critical for time-series data in a KD process [52].

## 5 CONCLUSION

In this work, we propose a novel Progressive Skeleton-to-sensor Knowledge Distillation (PSKD) model which only needs to accept time-series data *i.e.*, accelerometer data, from a smartwatch during the testing phase. Specifically, we propose the construction of multiple teacher models using both teacher and student modalities. In addition, we design an effective progressive learning scheme to eliminate the performance gap between the teacher and the student models. After that, a novel loss function called Adaptive-Confidence Semantic (ACS), is introduced to allow the student model adaptively to select the correct teacher model or the ground-truth label it needs to mimic. Extensive experimental results on UTD-MHAD, MMAct and Berkeley-MHAD datasets confirm the effectiveness and competitive performance compared to the previous methods on the mono sensor-based HAR problem.

## ACKNOWLEDGMENTS

This research is supported by NSF SCH-2123749 and SCH-2123521 Collaborative Research. This article solely reflects the opinions and conclusions of its authors and not the funding agents.